\documentclass[runningheads,a4paper]{llncs}

\usepackage{amssymb}
\usepackage{graphicx}
\usepackage{booktabs}
\usepackage{amsmath}
\usepackage[table,xcdraw]{xcolor} 
\usepackage{appendix}
\usepackage{url}
\urldef{\mailsa}\path|{arijit.sehanobish, nathaniel.brown, ishita.daga, jayashri.pawar,|
\urldef{\mailsb}\path|danielle.torres, anasuya.das, murray.becker, rherzog,|
\urldef{\mailsc}\path|benjamin.odry, ron.vianu}@coverahealth.com|    
\newcommand{\keywords}[1]{\par\addvspace\baselineskip
\noindent\keywordname\enspace\ignorespaces#1}
\graphicspath{ {./images/} }
\begin{document}

\mainmatter  

\title{Efficient Extraction of Pathologies from C-Spine Radiology Reports using Multi-Task Learning}

\titlerunning{Multi-Task Learning for Pathology Extraction}

%
%
\author{Arijit Sehanobish%
\thanks{Corresponding Author}%
\and Nathaniel Brown\and Ishita Daga\and Jayashri Pawar\and Danielle Torres\and Anasuya Das\and Murray Becker\and Richard Herzog\and Benjamin Odry\and Ron Vianu
}
\authorrunning{Multi-Task Learning for Pathology Extraction}

\institute{Covera Health,\\
NYC, New York\\
\mailsa\\
\mailsb\\
\mailsc\\
}

%
%

\toctitle{Lecture Notes in Computer Science}
\tocauthor{Authors' Instructions}
\maketitle

\begin{abstract}
Pretrained Transformer based models finetuned on domain specific corpora have changed the landscape of NLP. Generally, if one has multiple tasks on a given dataset, one may finetune different models or use task specific adapters. In this work, we show that a multi-task model can beat or achieve the performance of multiple BERT-based models finetuned on various tasks and various task specific adapter augmented BERT-based models. We validate our method on our internal radiologist's report dataset on cervical spine. We hypothesize that the tasks are semantically close and related and thus multitask learners are powerful classifiers. Our work opens the scope of using our method to radiologist's reports on various body parts. 
\end{abstract}

\section{Introduction}\label{sec:intro}

Since the seminal work by~\cite{vaswani2017attention}, Transformers have become the de-facto architecture for most Natural Language Processing (NLP) tasks. Self-supervised pretraining of massive language models like BERT~\cite{devlin2019bert} and GPT~\cite{brown2020language} has allowed practitioners to use these large language models with little or no finetuning to various downstream tasks. Multitask learning (MTL) in NLP has been a very promising approach and has shown to lead to performance gains even over task specific fine-tuned models~\cite{Worsham2020}. However, applying these large pre-trained Transformer models to downstream medical NLP tasks is still quite challenging. Medical NLP has its unique challenges ranging from domain specific corpora, noisy annotation labels and scarcity of high quality labeled data. In spite of these challenges, a number of different groups have successfully finetuned these large language models for various medical NLP tasks. However, there is not much literature that uses multi-task learning in medical NLP to classify and extract diagnoses in clinical text~\cite{peng2020empirical}. Moreover, there is almost no work in predicting spine pathologies from radiologists' notes~\cite{asj}. 

In this article, we are interested in extracting information from radiologists' notes on the cervical spine. In a given note, the radiologist discusses the specific, often multiple pathologies present on medical images and usually grade their severity. Extracting pathology information from a cervical spine report can facilitate the creation of structured databases that can be used for a number of downstream use-cases, such as cohort creation, quality assessment and outcome tracking. We focus on four of the most common pathologies in the cervical spine - central canal and foraminal stenosis, disc herniation and cord compression. Next, we create multiple tasks on a given report, where each task is to predict the severity of a pathology for each motion segment - the smallest physiological motion unit of the spinal cord~\cite{Swartz2005}. Breaking information down to the motion segment level enables any pathological findings to be correlated with clinical exam findings and could feasibly inform future treatment interventions. Given the semantic similarities between pathologies and the co-occurrence of multiple pathologies in a given sentence, we believe that these tasks are similar. Thus it is tempting to ask whether MTL approach can match the performance of task specific models, since it will cut down on the hardware requirements for training and will be faster during inference time. A number of different approaches have looked at task similarity and semantics to understand what tasks need to be grouped together or the conditions required for MTL to succeed~\cite{bingel-sogaard-2017-identifying},~\cite{zamir2020robust} and~\cite{standley2020tasks}.  Inspired by the theoretical results in~\cite{wass_mt}, we hypothesize that if the Wasserstein distance between tasks is small, a single multitasking model can match the performance of task specific models. 

In this work, we (\textbf{a}) design a novel pipeline to extract and predict the severity of various pathologies at \textit{each motion segment}, (\textbf{b}) quantify the notion of task similarity by computing the Wasserstein distance between tasks, and (\textbf{c}) show how to leverage that information into a simple MTL framework allowing us to achieve significant model compression during deployment and also speed up our inference without sacrificing the accuracy of our predictions. 

\begin{table}[h]
\centering
\resizebox{\columnwidth}{!}
{
\begin{tabular}{lcccc}
  Split   & Stenosis & Disc & Cord & Foraminal \\
  \midrule
   Train  &  \begin{tabular}[c]{@{}c@{}} None/Mild :  5488 \\ 
    Moderate : 561 \\
   Severe : 178
   \end{tabular}
   & \begin{tabular}[c]{@{}c@{}} None/Mild :  2731 \\ 
    Moderate : 2699 \\
   Severe : 797
   \end{tabular}& \begin{tabular}[c]{@{}c@{}} None :  5702 \\ 
    Mild/Severe : 525 
   \end{tabular} &  \begin{tabular}[c]{@{}c@{}} None :  5262 \\ 
    Severe : 965
   \end{tabular}\\  
  \midrule
  Test &  \begin{tabular}[c]{@{}c@{}} None/Mild :  793 \\ 
    Moderate : 68 \\
   Severe : 19
   \end{tabular}
   & \begin{tabular}[c]{@{}c@{}} None/Mild :  401 \\ 
    Moderate : 378 \\
   Severe : 101
   \end{tabular}& \begin{tabular}[c]{@{}c@{}} None :  806 \\ 
    Mild/Severe : 74 
   \end{tabular} &  \begin{tabular}[c]{@{}c@{}} None :  789 \\ 
    Severe : 91 
   \end{tabular}\\   
  \bottomrule
\end{tabular}
}
\caption{Statistics of our Dataset}
\label{tab:Dataset-stats}
\end{table}

\section{Datasets}
\begin{figure}[h]
	\centering
	\includegraphics[width=\columnwidth]{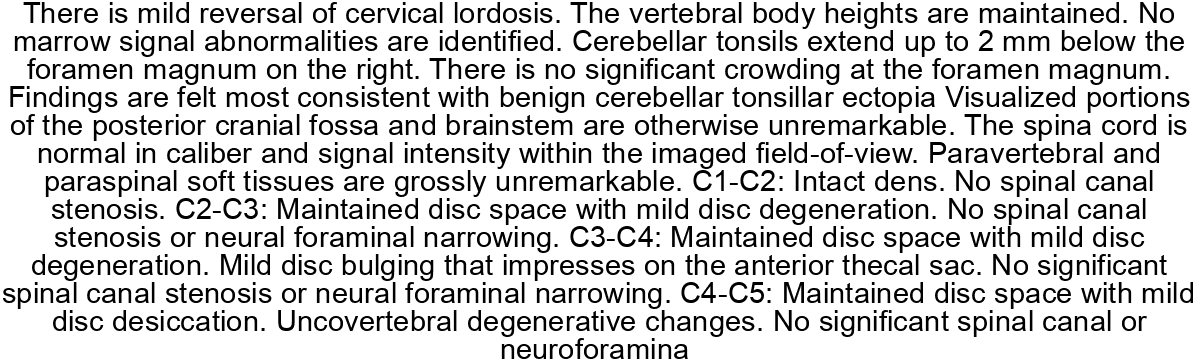}
	\caption{Example of our Dataset.}
	\label{Fig:Data-example}
\end{figure}

We use our internal dataset consisting of radiologists' MRI reports on cervical spine. The data consists of $1578$ reports coming from $97$ different radiology practices detailing various pathologies of the cervical spine. Our dataset is heterogeneous and is diversely sampled from a large number of different radiology practices and medical institutions. We annotate the data with the $4$ following pathologies : spinal stenosis, disc herniation, cord compression and neural foraminal stenosis. Each of these pathologies is bucketed into various severity categories. For central canal stenosis, the 3 categories are based on gradation; none/mild are not clinically significant, moderate and severe definitions involved cord compression or flattening. The moderate vs. severe gradation refers to the varying degrees of cord involvement. For disc herniation like central canal stenosis, the categories are based on a continuous spectrum and it is a standard practice in radiology for any continuous spectrum to be bucketed in mild, moderate and severe discrete categories. For cord compression, it is a binary classification problem : compression/signal change vs. none. This is because either cord compression or signal change can cause symptoms and is therefor clinically relevant. For foraminal stenosis, we are only interested in binary classification as well: severe vs non-severe as severe foraminal stenosis may indicate nerve impingement which is clinically significant. The splits and the details of each category can be found in Table~\ref{tab:Dataset-stats}. The data distribution is highly imbalanced and about $20\%$ of these reports are OCR-ed, which leads to additional challenges stemming from bad OCR errors. An example of a cervical report can be found in figure~\ref{Fig:Data-example}.  

\section{Description of the Workflow}
In this section, we will briefly describe our novel pipeline. The reports are first de-identified according to HIPAA regulations and the Spacy~\cite{spacy} parser is used to break the report into sentences. Then, each sentence is tagged by annotators and given labels of various pathologies and severities if the sentence mentions that pathology. For example, in the above report, the sentence: ``C1-C2: No significant neuroforaminal or spinal canal narrowing " : will be given normal or $0$ class for each of the $4$ pathologies. A BERT based NER model which we call the report segmenter is then used to identify the motion segment(s) present in a particular sentence and all the sentences containing a particular motion segment are concatenated together. The BERT based NER model achieves an F1 score of $.9$. More details about the NER model and the hyperparameters used to train it can be found in Appendix A. All pathologies are predicted using the concatenated text for a particular motion segment. Finally, the severities for each pathology are modeled as a multi-label classification problem, and a pre-trained transformer is finetuned using the text for each motion segment. For more details about our pipeline and data processing, please see Appendix B.
Figure~\ref{Fig:databreak} breaks down how a report looks as it goes through our pipeline. 
\begin{figure}[ht]
	\centering
	\includegraphics[width=.8\textwidth]{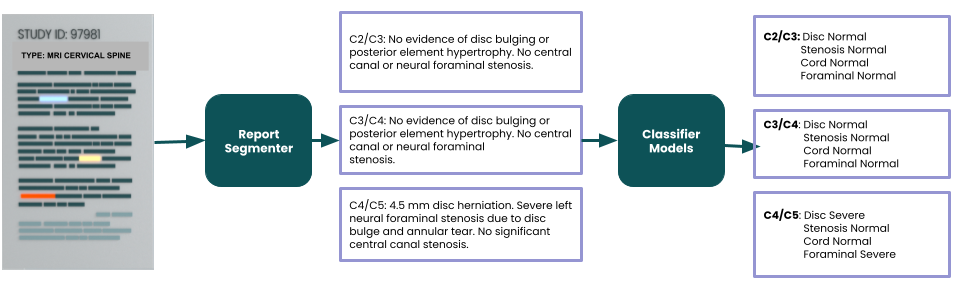}

	\caption{Figure showing how a report looks as it goes through our pipeline.}
	\label{Fig:databreak}
\end{figure}

\begin{table}[h]
\resizebox{\columnwidth}{!}{
\begin{tabular}{llcccc}
Backbone & Model & Stenosis & Disc & Cord & Foraminal \\
\midrule
& \begin{tabular}[c]{@{}c@{}}  Baseline\\ (single tasker) \end{tabular}  & $.62 \pm .03$ & $.64 \pm .03$ & $.70 \pm .03$ & $.79 \pm .03$ \\
 & MultiTasking & $.62 \pm .02$ & $.65 \pm .03$ & $.72 \pm .02$ & $.78 \pm .01 $\\
\begin{tabular}[c]{@{}c@{}} BERT \\ BASE \end{tabular} & \begin{tabular}[c]{@{}c@{}} 
   Task specific \\ Adapters \end{tabular} & $.63 \pm .01$ & $.64 \pm .03$ & $.68 \pm .02$ & $.79 \pm .02$ \\
 & \begin{tabular}[c]{@{}c@{}} Multitasking \\ Adapter \end{tabular} & $.65 \pm .02$ & $.66 \pm .03$ & $.70 \pm .04$ & $.80 \pm .03$ \\
 & \begin{tabular}[c]{@{}c@{}} Fusion Adapters \\ \cite{pfeiffer2020AdapterHub} \end{tabular}  & $.64 \pm .03$ & $.66 \pm .01$ & $.70 \pm .03$ & $.79 \pm .03$ \\
 \midrule 
& \begin{tabular}[c]{@{}c@{}}  Baseline\\ (single tasker) \end{tabular}  & $.64 \pm .05$ & $.66 \pm .02$ & $.71 \pm .02$ & $\mathbf{.82 \pm .01}$ \\
 & MultiTasking & $.63 \pm .02$ & $\mathbf{.67 \pm .01}$ & $\mathbf{.75 \pm .01}$ & $.79 \pm .03 $\\
\begin{tabular}[c]{@{}c@{}} CLINICAL \\ BERT \end{tabular} & \begin{tabular}[c]{@{}c@{}} 
   Task specific \\ Adapters \end{tabular} & $\mathbf{.66 \pm .01}$ & $.65 \pm .03$ & $.69 \pm .03$ & $.81 \pm .01$ \\
 & \begin{tabular}[c]{@{}c@{}} Multitasking \\ Adapter \end{tabular} & $\mathbf{.66 \pm .01}$ & $\mathbf{.67 \pm .03}$ & $.71 \pm .03$ & $.81 \pm .02$ \\
 & \begin{tabular}[c]{@{}c@{}} Fusion Adapters \\ \cite{pfeiffer2020AdapterHub} \end{tabular}  & $.65 \pm .03$ & $\mathbf{.67 \pm .01}$ & $.72 \pm .02$ & $.81 \pm .02$ \\

\bottomrule
\end{tabular}
}
\caption{Table showing the macro F1 scores over 5 trials of our Baseline and  MultiTasking Models}
\label{tab:Expt-results}
\end{table}

\section{Methods}
Two types of multitasking models are assessed: (\textbf{a}) A multitasking BERT model and (\textbf{b}) A multitasking adapter augmented BERT model. We consider Adapters in our experiments as they provide a simple way to quickly train these transformer models on a limited computational budget and on smaller datasets.

For our classification tasks, Clinical BERT model~\cite{alsentzer2019publicly} is used as a backbone. The Clinical BERT model is then finetuned on the above tasks resulting in $4$-task specific BERT sequence classifier models which provide our baseline results. 

Now, instead of finetuning $4$-BERT based models,  $4$ classifier heads (i.e. $4$ linear layers) is applied to a single Clinical BERT model to create an output layer of shape $[3,3,2,2]$, where the first $3$-outputs correspond to the logits for the stenosis severity prediction, the next $3$ for the disc severity, the next $2$ for the cord severity and the final $2$ logits for the foraminal severity. A dropout of $.5$ is added to the BERT vectors before passing them to the classifier layers.  Each of these classifier heads is trained with a cross entropy loss with the predicted logits and the ground truth targets. All the losses are added as in the equation below~\ref{eqn:Joint loss} which allows the gradients to back propagate through the whole model and train these classifier heads jointly. 
\begin{equation*}\label{eqn:Joint loss}
\mathcal{L} = l_{\text{stenosis}} + l_{\text{disc}} + l_{\text{cord}} + l_{\text{foraminal}}
\end{equation*}
where $l_{\text{pathology}}$ is the cross-entropy between the predicted logits for the given pathology severity and the ground truth labels.  

Finetuning these large transformer models is expensive, and sometimes they do not show much improvement where there is a lack of training data. To alleviate these problems~\cite{houlsby2019parameterefficient} and~\cite{pfeiffer2020AdapterHub} introduce a novel parameter efficient transfer and multitask learning technique by adding in small networks called Adapters in between various Transformer blocks. Adapter modules perform more general architectural modifications to re-purpose a pretrained network for a downstream task. In standard fine-tuning, the new top-layer and the original weights are updated. In contrast, in Adapter tuning, the parameters of the original network are frozen and therefore may be shared by many tasks. Given the success of adapters for MTL, we experiment by adding Adapters as described in~\cite{pfeiffer2021adapterfusion} in between every BERT output layers. When training with Adapters, the BERT weights are frozen. For multitasking,~\cite{pfeiffer2021adapterfusion} proposed splitting and fusion adapters to prevent catastrophic forgetting across tasks. But in our work, the same adapters across all tasks are used like~\cite{stickland2019bert}. But unlike the above work, our results match the results with task-specific adapters or the fusion adapters for multitask learning. Our implementation follows the one outlined in~\cite{pfeiffer2020AdapterHub} and thus our work can be seen as a simplification of the training strategies as proposed above without sacrificing the accuracy or speed. We conjecture that the tasks are similar and the sentences across these tasks have similar structure and semantic meaning, which allows these multitask models to perform without any need for task specific architectures. These hypotheses are validated by computing Wasserstein distances between various tasks in Section~\ref{wass}. PyTorch and the Hugging Face Library~\cite{wolfetal2020transformers} is used to train our models on NVIDIA V100 16GB GPU and the POT library~\cite{flamary2021pot} is used to compute the Wasserstein distances. More training details can be found in the Appendix A.

\section{Results}
In this section, we validate our multitasking models on our cervical dataset. For detailed comparison, we also experiment with our multitasking and Adapter based models starting with the weights of BERT-base. Table~\ref{tab:Expt-results} shows the results of our multitasking models over our baseline models and the adapter augmented models. Models which are initialized with the weights of Clinical BERT show an improvement over the corresponding models initialized with the weights of BERT base. Moreover, our results show that the multitasking models perform as well as the task specific models. In fact, our results with the Fusion adapter modules show that mixing information from various tasks can actually improve model performance. Finally, table~\ref{tab:walltimes} shows significant improvements in inference speeds on our test set of the multitasking models over the baseline single taskers.
\begin{table}
    \centering
    
\resizebox{\columnwidth}{!}{\begin{tabular}{lcccc}
      Model   & \begin{tabular}[c]{@{}c@{}}  Baseline Clinical BERT \\ (single tasker) \end{tabular}  & Multitasking Clinical BERT & \begin{tabular}[c]{@{}c@{}} Baseline Clinical BERT with \\
   Task specific Adapters \end{tabular} & \begin{tabular}[c]{@{}c@{}} Multitasking Adapter \\
   Augmented Clinical BERT \end{tabular}\\
   \midrule
      Walltime (seconds)   & $259.64$ & $\mathbf{56.93}$ & $281.56$ & $\mathbf{60.16}$\\
      \bottomrule
    \end{tabular}
    }
    \caption{Table showing faster inference speed of our Multitasking models over the Baseline Models}
    \label{tab:walltimes}
\end{table}

\begin{table}
\resizebox{\columnwidth}{!}{
\begin{tabular}{lccccc}
Task  &  Mild Stenosis & Severe Disc & Mild Disc & Mild/Severe Foraminal & Mild/Severe Cord\\ \midrule
Mild Stenosis&
  0 &
  \cellcolor[HTML]{F8FF00}$.7 \pm .4$ &
  \cellcolor[HTML]{FFFC9E}$.3 \pm .2$ &
  \cellcolor[HTML]{FFCC67}$1.2 \pm .3$ &
  \cellcolor[HTML]{F8FF00}$.7 \pm .6$ \\
Severe Disc &
  \cellcolor[HTML]{F8FF00}$.7 \pm .4$ &
  0 &
  \cellcolor[HTML]{FFFFC7}$.2 \pm .1$ &
  \cellcolor[HTML]{FFD788}$.8 \pm .6$ &
  \cellcolor[HTML]{FEFD73}$.6 \pm .5$ \\
Mild Disc &
  \cellcolor[HTML]{FFFC9E}$.3 \pm .2$ &
  \cellcolor[HTML]{FFFFC7}$.2 \pm .1$ &
  0 &
  \cellcolor[HTML]{FCFF2F}$.7 \pm .5$ &
  \cellcolor[HTML]{FFCB2F}$1.1 \pm .7$ \\
Mild/Severe Foraminal &
  \cellcolor[HTML]{FFCC67}$1.2 \pm .3$ &
  \cellcolor[HTML]{FFD788}$.8 \pm .6$ &
  \cellcolor[HTML]{FCFF2F}$.7 \pm .5$ &
  0 &
  \cellcolor[HTML]{FFD37D}$.8 \pm .7$ \\
Mild/Severe Cord &
  \cellcolor[HTML]{F8FF00}$.7 \pm .6$ &
  \cellcolor[HTML]{FEFD73}$.6 \pm .5$ &
  \cellcolor[HTML]{FFCC67}$1.1 \pm .7$ &
  \cellcolor[HTML]{FFD37D}$.8 \pm .7$ &
  0 \\ \hline
\end{tabular}
}
\caption{Table showing the Sliced Wasserstein Distance between Tasks on the Training Set. Some of the Wasserstein distances are not shown as they can not be computed owing to the sample size of some of our minority classes.}
   \label{tab:Wasserstein}

\end{table}

\section{Empirical Evidence behind MultiTasking Models}\label{wass}
In this section, we provide some evidence behind the performance of our MultiTasking models. There is an explicit relationship between the Wasserstein distances and the generalization error in MTL as proposed in Theorem 1 and Theorem 2 in~\cite{wass_mt}. Moreover, motivated by the work of~\cite{alvarezmelis2020geometric} to compute distances between labeled datasets, we define a task $\mathcal{T}_y(X) := P(X | Y = y)$ as a conditional distribution. We then define the distance between the tasks to be :
\begin{equation}~\label{eqn:dist}
d((X, y),(X', y')) := W_{2}(\mathcal{T}_y(X), \mathcal{T}_{y'}(X') )
\end{equation}
where $W_2$ is $2$-Wasserstein distance. This conditional distribution also appears in the work of~\cite{courty2017joint}. Computing Wasserstein distances are extremely computationally expensive. Thus, various authors approximate $W_2$ by the Wasserstein-Bures metric or by various entropic regularized Sinkhorn divergences~\cite{alvarezmelis2020geometric} and~\cite{chizat2020faster}. 

Here, $W_2$ metric is approximated by the sliced Wasserstein distance~\cite{kolouri2019generalized} with $60$ random projections of dimensions in logspace between $1$ and $4$. To apply the sliced Wasserstein distance, the embeddings from the final BERT layer of our pretrained task specific BERT models are extracted, i.e. $X$ is the $768$-dimensional vector representation of the [CLS] token coming from the appropriate BERT model. The sliced Wasserstein distances are computed between these BERT embeddings. Table~\ref{tab:Wasserstein} shows the sliced Wasserstein distances between various conditional distributions. The upper bound for Wasserstein distance between two probability measures is given by:
\begin{equation*}\label{eqn:ub}
    W_{2}(\mathcal{T}_y(X), \mathcal{T}_{y'}(X') ) \leq diam(A)TV(\mathcal{T}_y(X), \mathcal{T}_{y'}(X') )
\end{equation*}
where diam(A) is the diameter of the support of the measures and in our cases can be bounded by $59.4$, similar to the reported values in~\cite{kobayashi2020attention}. $TV(\mathcal{T}_y(X), \mathcal{T}_{y'}(X')$) is the total variation and can be trivially bounded by 1. 
The  relatively small  distances (which are also considerably lower than the upper bound)  between  tasks  is  most  likely  why  a multitask model is able to replicate the performances of task specific models. The bound given by the above equation is not tight so we provide some empirical analysis on the bounds of the Wasserstein distances by using the methods of~\cite{alvarezmelis2020geometric} on some public text classification datasets. This analysis can be found in Appendix C.

\section{Conclusion}
In this work, a simple multitasking model is presented that is competitive with task specific models and is $4$ times faster during inference time. Instead of training and deploying $4$ models, only one model is trained and deployed, thus achieving significant model compression. This work opens the possibility of using multitasking models to extract information over various different body parts and thus allows users to leverage these large transformer models using limited resources. Our novel pipeline is one of the very few works that attempts to extract pathologies and their severities from a heterogeneous source of radiologists' notes on cervical spine MRIs at the level of \textit{motion segments}. In these ways, our findings suggest that our approach may not only be more widely generalizable and applicable but also more clinically actionable.  Finally, we also shed light on how closely related or semantically similar these tasks are. In our future work, we will expand on our observations to the radiologist's reports for other body parts and other pathologies for the cervical spine. We will also focus on further characterizing medical NLP datasets and tasks using our definition of task similarity so we can define when learning can be cooperative and when learning is competitive and whether our definition of task similarity has any clinical significance.

\bibliographystyle{unsrt}
\bibliography{ref.bib}

\appendix

\begin{figure*}[ht]
	\centering
	\includegraphics[width=.9\textwidth]{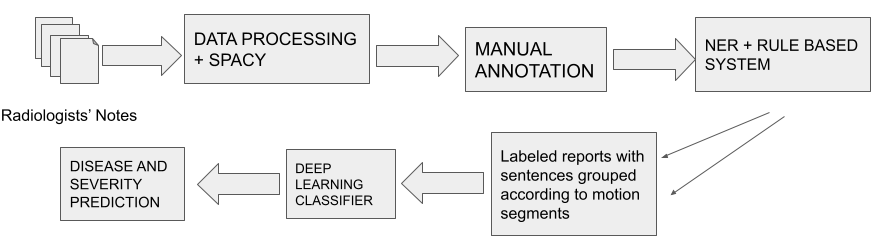}

	\caption{Workflow of Our Approach}
	\label{Fig:workflow}
\end{figure*}
\section{Training Details}
We create a validation set using $10\%$ of the samples of the training set where the samples are drawn via stratified samples so the data distribution is maintained across splits. For finetuning the BERT models (BERT-Base and Clinical BERT) or the multitasking model, we finetune the whole model with a batch size of $16$ for $3-5$ epochs with BERT Adam optimizer which is basically a weight decoupled Adam optimizer~\cite{adamw}.  The learning rate used is $1e-5$ with a linear learning rate decay scheduler and the weight decay is $1e-4$. For training the Adapter augmented models, we use a higher learning of $1e-4$. We use the Adapter architecture as defined in~\cite{pfeiffer2020AdapterHub}. Adapters are 2 layer feedforward network with a bottle neck dimension of $48$. We initialize the adapter weights such that initially the whole Adapter layer is almost an identity function. Thus one can think of these adapter architectures as autoencoder architectures. We experimented with both GELU and ReLU as non-linearities between the feedforward layers and the results were similar. We train our adapter augmented models for about $10-12$ epochs with early stopping on the validation loss. For Adapter augmented models, the BERT weights were frozen and only the adapter and classifier weights were updated. All the baseline models were finetuned for 15 epochs and the multitasking model was finetuned for 12 epochs. The Adapter based models were trained for 20 epochs and the Multitasking Adapter model is trained for 23 epochs. The sequence length used for the all the baseline and the multitasking classifier models is $512$. 

The NER model is a BERT-based binary classifier (Location Tag vs the Other Tag). It is our in-house model that is trained on both lumbar and cervical MRI reports (about $6000$ reports) that can predict the location tags in those reports. The model is trained for $5$ epochs with a batch size of $16$ and sequence length $256$. We used AdamW optimizer with weight decay of $1e-4$. The learning rate used is $1e-5$ with a linear learning rate decay scheduler. Our NER model achieves an F1 score of $.9$.

\section{Detailed Description of our Workflow}
In this section, we give a more detailed description of our workflow. Our main goal is to detect pathologies at the \textit{motion segment} level from radiologists' MRI reports on cervical spine.  The motion segments we care about in our work are C2-C3, C3-C4, C4-C5, C5-C6, C6-C7 and C7-T1. We first make sure that the reports are de-identified and then use a Spacy~\cite{spacy} parser to break the report into sentences. Then each sentence is tagged by annotators and they are given labels of various pathologies and their severities if the sentence mentions that pathology. To detect pathologies at a motion segment level, we use our BERT based NER system to tag the locations present in each sentence. Our tag of interest for the NER model is the motion segment tag and we achieve an F1 score of $.9$ for that tag. We then use a rule based system to group all sentences to the correct motion segment. If a sentence does not explicitly have a motion segment mentioned it, we use a rule based method to assign the sentence to one of the above mentioned motion segments or to a generic category ''No motion segments found". Given the disparate source of our data, for example, C34, C3-C4, $\text{C3}\_$C4 all refer to the motion segment C3-C4 and thus our systems are mindful of this diversity of the clinical notes. Finally to use our BERT based models for pathology detection on the level of motion segments for a given report, we concatenate all sentences for a given motion segment and use the [CLS] token for the segment that is used for the downstream classification task.  Figure~\ref{Fig:workflow} shows our workflow. 

Since we are interested in predictions at the motion segment level, we do not use the sentences that are grouped under ''No segment found" to train the classifier models, nor do we evaluate our classifier models on those sentences.

\section{Bounds on Wasserstein distances between Text Datasets}

The upper bound for the Wasserstein distance derived in our paper is not tight. We are unable to provide a tighter bound without additional hypothesis on the empirical distributions considered in our work. 
Instead we provide clarity on how large these numbers can be between various publicly available datasets. The following comparisons are inspired by the results described in Figure 8 in~\cite{alvarezmelis2020geometric}. The distance between labeled datasets as described in the above paper has two components: \textbf{(1)} Euclidean distance between the feature vectors and \textbf{(2)} Wasserstein distance between the conditional distributions (also described in~\cite{courty2017joint}). Disentangling the contribution of the feature vectors following the method outlined in their appendix, we find the Wasserstein distance between positive labels in the Yelp binary polarity dataset and the ``Educational Institution" class in the DbPedia-14 to be as large as $37.67$ and distances as low as $.64$ between positive classes in Amazon reviews binary polarity and the Yelp binary polarity datasets. We hope that these numbers shed some light on the possible range of the Wasserstein distances between some benchmark text classification datasets. However, given the SOTA performances by BERT on these datasets is almost perfect and the possibility of DbPedia leaking into BERT's training data (DbPedia is scrawled from Wikipedia), an MTL experiment on these datasets may not justify our hypothesis.  

\end{document}